\documentclass{article}

\usepackage{PRIMEarxiv}

\usepackage[utf8]{inputenc} 
\usepackage[T1]{fontenc}    
\usepackage{hyperref}       
\usepackage{url}            
\usepackage{booktabs}       
\usepackage{amsfonts}       
\usepackage{nicefrac}       
\usepackage{microtype}      
\usepackage{lipsum}
\usepackage{fancyhdr}       
\usepackage{graphicx}       
\graphicspath{{media/}}     

\usepackage{perpage}
\MakePerPage{footnote}
\usepackage{comment}
\usepackage{paralist}
\usepackage{bbding}

\usepackage{hyperref}       
\usepackage{url}            
\usepackage{booktabs}       
\usepackage{amsfonts}       
\usepackage{nicefrac}       
\usepackage{microtype}      
\usepackage{xcolor}         

\usepackage[table]{xcolor}  
\definecolor{perfLow}{HTML}{FFF5F5}
\definecolor{perfHigh}{HTML}{E98C8C}

\ExplSyntaxOn
\prop_new:N \g_perf_data_prop
\prop_new:N \g_perf_min_prop
\prop_new:N \g_perf_max_prop
\seq_new:N \l_perf_parts_seq
\tl_new:N \l_perf_value_tl
\tl_new:N \l_perf_mean_tl
\tl_new:N \l_perf_min_tl
\tl_new:N \l_perf_max_tl
\fp_new:N \l_perf_score_fp
\cs_generate_variant:Nn \prop_gput:Nnn { Nnx }
\cs_new_protected:Npn \perf_register:nnn #1#2#3
  {
    \prop_gput:Nnn \g_perf_data_prop {#1/#2} {#3}
    \str_if_eq:nnF {#3} {--}
      {
        \seq_set_split:Nnn \l_perf_parts_seq {/} {#3}
        \tl_set:Nx \l_perf_mean_tl {\seq_item:Nn \l_perf_parts_seq {1}}
        \prop_get:NnNTF \g_perf_min_prop {#2} \l_perf_min_tl
          {
            \prop_get:NnN \g_perf_max_prop {#2} \l_perf_max_tl
            \prop_gput:Nnx \g_perf_min_prop {#2}
              {\fp_eval:n {min(\l_perf_min_tl,\l_perf_mean_tl)}}
            \prop_gput:Nnx \g_perf_max_prop {#2}
              {\fp_eval:n {max(\l_perf_max_tl,\l_perf_mean_tl)}}
          }
          {
            \prop_gput:Nnx \g_perf_min_prop {#2} {\l_perf_mean_tl}
            \prop_gput:Nnx \g_perf_max_prop {#2} {\l_perf_mean_tl}
          }
      }
  }
\NewDocumentCommand{\PerfDeclareRow}{mmmmmm}
  {
    \perf_register:nnn {#1} {params} {#2}
    \perf_register:nnn {#1} {ant} {#3}
    \perf_register:nnn {#1} {cube} {#4}
    \perf_register:nnn {#1} {fourroom} {#5}
    \perf_register:nnn {#1} {pusht} {#6}
  }
\NewDocumentCommand{\PerfCell}{mm}
  {
    \prop_get:NnN \g_perf_data_prop {#1/#2} \l_perf_value_tl
    \tl_if_eq:NnTF \l_perf_value_tl {--}
      {\texttt{--}}
      {
        \seq_set_split:NnV \l_perf_parts_seq {/} \l_perf_value_tl
        \tl_set:Nx \l_perf_mean_tl {\seq_item:Nn \l_perf_parts_seq {1}}
        \prop_get:NnN \g_perf_min_prop {#2} \l_perf_min_tl
        \prop_get:NnN \g_perf_max_prop {#2} \l_perf_max_tl
        \fp_compare:nNnTF {\l_perf_max_tl} = {\l_perf_min_tl}
          {\fp_set:Nn \l_perf_score_fp {0.5}}
          {
            \fp_set:Nn \l_perf_score_fp
              {(\l_perf_mean_tl-\l_perf_min_tl)/(\l_perf_max_tl-\l_perf_min_tl)}
            \str_if_eq:nnT {#2} {params}
              {\fp_set:Nn \l_perf_score_fp {1-\l_perf_score_fp}}
          }
        \exp_args:Nx \cellcolor
          {perfHigh!\fp_eval:n {100*\l_perf_score_fp}!perfLow}
        \str_if_eq:nnTF {#2} {params}
          {\l_perf_mean_tl}
          {\ensuremath{\l_perf_mean_tl\pm\seq_item:Nn \l_perf_parts_seq {2}}}
      }
  }
\ExplSyntaxOff

\usepackage{makecell}

\newcommand{\perfcolor}[2]{{\setlength{\fboxsep}{1pt}\colorbox{#1}{\strut #2}}}

\usepackage{amsmath,amssymb,amsthm}
\usepackage{algorithm,algorithmic}
\usepackage{booktabs}
\usepackage{graphicx}
\usepackage{caption}
\usepackage{subcaption}
\usepackage{enumitem}
\usepackage{mathtools}
\usepackage{bm}  
\usepackage{multirow}
\usepackage{comment}
\usepackage{wrapfig}

\title{Time-to-Goal: Learned Temporal Metrics for Long-Horizon Planning with JEPA World Models 
}

\author{
  Ruilong Yu \\
  University of Electronic Science\\ and Technology of China \\
  \texttt{yrl666@outlook.com} \\
   \And
  Fei Ye \thanks{denotes the corresponding author.} \\
  University of Electronic Science\\ and Technology of China \\
  \texttt{feiye@uestc.edu.cn} \\
}

\begin{document}
\maketitle

\begin{abstract}
Joint-Embedding Predictive Architectures (JEPAs) support planning by comparing predicted latent states with an encoded goal. For this comparison to guide temporal progress, states closer in time to the goal should receive lower costs. However, training representations for prediction and distribution regularization does not directly enforce this ordering. We examine this mismatch in flat LeWorldModel (LeWM) by analyzing how its Euclidean goal cost changes with temporal separation. Our analysis identifies conditions under which states at different long offsets receive similar average costs, weakening their temporal distinction. To learn the missing ordering explicitly, we introduce Time-to-Goal (T2G), a lightweight metric trained on latent triples from offline trajectories. A margin ranking loss encourages smaller distances for temporally nearer states, requiring no rewards or policy labels. The learned metric then replaces the Euclidean planning cost while the JEPA encoder and predictor remain frozen. Diagnostics on AntMaze and FourRoom show that T2G preserves stronger temporal contrast and improves near--far ranking at long offsets. Across four navigation and manipulation benchmarks, T2G improves the average planning success rate from 32.7\% (flat LeWM) to 63.0\% and approaches or even surpasses the performance of the hierarchical H-JEPA baselines. These results show that learning temporal ordering for the goal cost can substantially improve latent planning even with a flat world model. Code is available at \texttt{\href{https://github.com/yrluestc/T2G-JEPA}{https://github.com/yrluestc/T2G-JEPA}}.
\end{abstract}


\section{Introduction}
\begin{wrapfigure}{r}{0.50\linewidth}
\vspace{-\intextsep}
	\centering
	\includegraphics[width=0.99\linewidth]{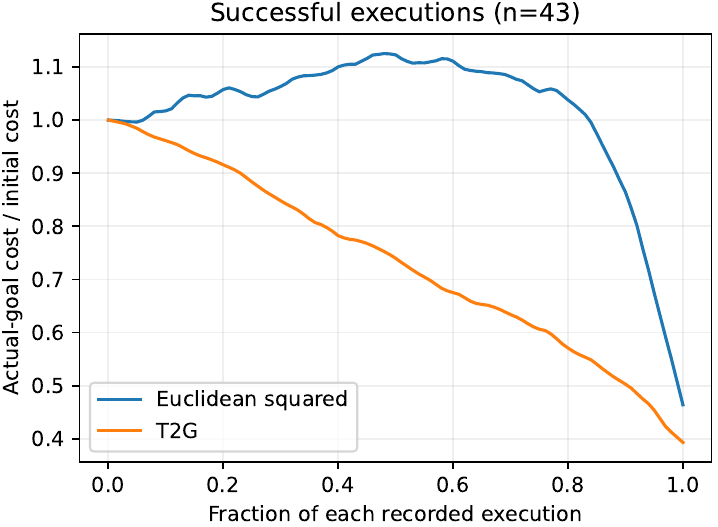}
\vspace{-5pt}
\caption{On successful FourRoom executions, Euclidean cost weakly signals initial progress (i.e., distant to goal), whereas T2G provides a more consistent temporal cost.
\vspace{-\intextsep}
}
\label{fig:plan_progress_fourroom}
\end{wrapfigure}

Joint-Embedding Predictive Architectures (JEPAs)~\cite{lecun2022path,assran2023self,bardes2024revisiting} learn world models by predicting future observations in a latent space. An encoder maps observations to representations, and an action-conditioned predictor models how these representations evolve under actions. This avoids pixel reconstruction and allows planning by comparing predicted future states with an encoded goal. LeWorldModel (LeWM)~\cite{LeWM} implements this with a next-embedding prediction loss and a sketched isotropic Gaussian regularizer (SIGReg), which discourages representation collapse. Its planner optimizes actions using the Euclidean distance between predicted states and the goal. However, as shown in Fig.~\ref{fig:plan_progress_fourroom}, this flat planner may provide an unreliable progress signal over long horizons, where reliable predictions and a cost that reflects distance to or progress toward a distant goal are both required. A flat model must roll out many prediction steps, leading to error accumulation. Moreover, matching a distant goal in a representation space that also carries low-level details can be difficult. H-JEPA~\cite{H-JEPA} addresses these challenges with a hierarchy of JEPAs operating at increasing temporal strides in which   subgoals divide a long task into shorter planning problems, and higher levels provide more abstract spaces for goal matching. Because this hierarchy changes both how futures are predicted and how they are evaluated, its gains conflate architectural and cost-related effects. This motivates isolating the goal cost and examining whether a better cost estimate alone can improve flat planning in long-horizon settings.

We ask whether the goal-matching cost of a flat model can guide planning without introducing a hierarchy. Intuitively, a useful cost should assign lower values to states that are closer to reaching the goal. However, LeWM’s Euclidean cost is applied to representations trained for next-state prediction and distribution regularization. Its training objective does not explicit impose supervision of progress toward a goal. We therefore examine the Euclidean costs of states along a saved planning trajectory. Specifically, we compare pairs of states separated by an interval of several model steps. Our analysis shows that, as this interval increases, the difference in their Euclidean costs becomes progressively smaller. This weakens the cost signal for ranking candidate futures when the states are far from the goal.

To provide this temporal signal, we propose \textbf{Time-to-Goal (T2G)}, a lightweight learned temporal metric for long-horizon planning. T2G learns a small embedding network on flat LeWM latents and uses Euclidean distance in this embedding as a temporal-distance score. It is trained on latent triples sampled from offline trajectories, where a ranking objective encourages a temporally nearer future state to be closer to the anchor state than a farther future state. The LeWM encoder and predictor remain \textit{frozen}, and no rewards or policy labels are required during this process. During planning, the same predictor rolls out candidate action sequences, and T2G scores their predicted states by their learned distance to the goal.

Our contributions are summarized as follows.
\begin{compactitem}
\item We analyze the case where Euclidean latent distances provide little distinction across temporal separations when states are far from the goal.
\item We introduce Time-to-Goal, a lightweight learned temporal metric with only 0.182M parameters (1.6\% of the flat LeWM), which substantially improves flat LeWM on long-horizon tasks without changing the latent world model. 
\item Comprehensive experimental results demonstrate that T2G enables flat LeWM to approach or even surpass multi-level H-JEPA in our long-horizon evaluations.
\end{compactitem}

\section{Methodology}
\label{sec:methodology}

\subsection{Problem Setup}
We consider an offline dataset of trajectories of length \(T\), expressed as:
\begin{equation}
	\mathcal{D}=\{(o_1,a_1,o_2,a_2,\ldots,o_T)\}.
	\label{eq:dataset}
\end{equation}
A frozen JEPA world model consists of an encoder \(E\), which maps observations to latents \({\bf z}_t=E(o_t)\in\mathbb{R}^{D}\), and an action-conditioned predictor \(F\), where \(D\) is the latent dimension. At each replanning call, we index the current latent as \({\bf z}_1\) and encode the goal observation as \({\bf z}_g=E(o_g)\). The planner samples or optimizes candidate action sequences \(a_{1:H}\), where \(H\) is the planning horizon. Each candidate produces a latent rollout \({\hat {\bf z}}_{2:H+1}\) through
\begin{equation}
    {\hat {\bf z}}_{t+1}=F(\hat {\bf z}_t,a_t),\qquad \hat {\bf z}_1={\bf z}_1,
    \label{eq:predictor}
\end{equation}
for \(t=\{1,\ldots,H\}\). The planner selects the action sequence with the lowest goal cost. In the H-JEPA implementation of flat LeWM, this cost is computed over the last-\(n\) predicted latents with \(n=\min(n_{\mathrm{last}},H)\):
\begin{equation}
\begin{aligned}
	C_{\mathrm{flat}}({\hat {\bf z}}_{2:H+1})=\sum\nolimits_{t=H-n+2}^{H+1}\lVert \hat {\bf z}_t-{\bf z}_g\rVert_2^2,
	\label{eq:flat_cost}
\end{aligned}
\end{equation}
where \(n_{\mathrm{last}}\) is the predefined cost window and $||\cdot||^2_2$ denotes the L2 norm. We seek a replacement cost that directly reflects relative progress toward the goal while keeping the encoder and predictor fixed.

\subsection{Why Euclidean Distance May Provide Weak Temporal Guidance}
\label{sec:euclidean_cost}
The cost in Eq.~\eqref{eq:flat_cost} guides action search by preferring predicted states with smaller squared Euclidean distances to the goal. To reflect temporal progress, this cost should generally assign lower values to states that are fewer steps from the same goal along a trajectory. LeWM trains its encoder and predictor with a next-embedding prediction loss and SIGReg:
\begin{equation}
	\mathcal{L}_{\mathrm{pred}}=\lVert {\hat {\bf z}}_{t+1}-{\bf z}_{t+1}\rVert_2^2,
	\qquad
\mathcal{L}_{\mathrm{LeWM}}=\mathcal{L}_{\mathrm{pred}}+\lambda\,\mathrm{SIGReg}(Z),
	\label{eq:lewm_loss}
\end{equation}
where \({\hat {\bf z}}_{t+1}=F({\bf z}_{t-H+1:t},a_{t-H+1:t})\), \(Z\in\mathbb{R}^{N\times D}\) collects \(N\) latent embeddings, and \(\lambda>0\). SIGReg projects the embeddings onto \(M\) random unit directions \(u^{(m)}\) on the unit sphere \(\mathcal{S}^{D-1}\subset\mathbb{R}^{D}\), and matches each one-dimensional projection to a standard Gaussian with an Epps--Pulley~\cite{epps1983test} statistic \(T_{\mathrm{EP}}\):
\begin{equation}
\begin{aligned}
	\mathrm{SIGReg}(Z)=\frac{1}{M}\sum\nolimits_{m=1}^{M}T_{\mathrm{EP}}(Zu^{(m)}).
	\label{eq:sigreg}
    \end{aligned}
\end{equation}

The prediction loss encourages accurate latent dynamics, while SIGReg controls the embedding distribution to discourage collapse. Neither objective explicitly requires a state closer in time to a goal to have a smaller distance to it. To understand the resulting cost signal, we examine how the average squared Euclidean distance changes with temporal separation. Let \((X_k,Y_k)\) denote an earlier encoded state and a future goal sampled from the same trajectory, separated by \(k\) model steps. Write the means of the paired states as \(\mu_X(k),\mu_Y(k)\), their covariances as \(\Sigma_X(k),\Sigma_Y(k)\), and their cross-covariance as \(C(k)\). For finite second moments, expanding the squared distance yields:
\begin{equation}
	\begin{aligned}
	D_E(k)&:=\mathbb{E}\lVert X_k-Y_k\rVert_2^2\\
	&=\lVert\mu_X(k)-\mu_Y(k)\rVert_2^2
	+\operatorname{tr}(\Sigma_X(k))+\operatorname{tr}(\Sigma_Y(k))
	-2\operatorname{tr}(C(k))\\
	&=:B(k)-2q(k),
	\end{aligned}
	\label{eq:general_latent_distance}
\end{equation}
where \(B(k)\) collects the squared mean difference and the two marginal variance traces, and \(q(k)=\operatorname{tr}(C(k))\). Thus, the average distance depends on both the distributions of the individual states and how the paired states vary together.

For planning, the relevant question is whether this distance distinguishes states that are closer to the same future goal. Holding the sampled future goals fixed, we compare states at offsets \(k\) and \(k+\Delta\), where \(\Delta>0\). The difference in their average costs, which we call the \emph{temporal contrast}, is
\begin{equation}
	\begin{aligned}
	S_E(k,\Delta):=D_E(k+\Delta)-D_E(k)=\delta_\Delta B(k)-2\delta_\Delta q(k),
	\end{aligned}
	\label{eq:temporal_distance_contrast}
\end{equation}
where \(\delta_\Delta f(k)=f(k+\Delta)-f(k)\). Positive contrast means that the temporally farther states receive higher costs on average. The triangle inequality gives:
\begin{equation}
	|S_E(k,\Delta)|\le|\delta_\Delta B(k)|+2|\delta_\Delta q(k)|.
	\label{eq:temporal_contrast_bound}
\end{equation}
This bound explains why Euclidean cost can provide weak temporal guidance at large \(k\): if \(B(k)\) and \(q(k)\) change little as the offset increases from \(k\) to \(k+\Delta\), states at these two offsets from the same goal receive similar average costs, making their temporal proximity less distinguishable through Euclidean cost.

\begin{figure*}[t]
\centering
\includegraphics[width=0.99\linewidth]{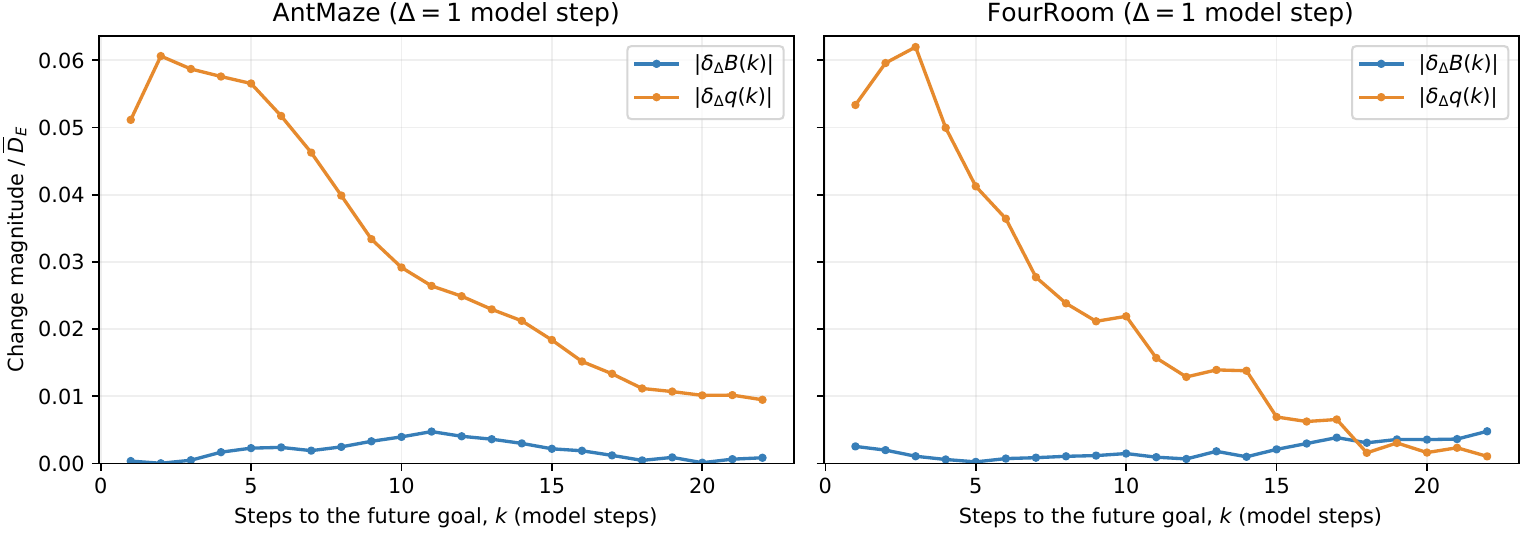}
\caption{Changes in the components of Euclidean temporal contrast on AntMaze (left) and FourRoom (right), using 50 recorded planner executions per environment. The horizontal axis is the separation \(k\) from a sampled future goal in model steps, with a fixed comparison gap of \(\Delta=1\) model step (five environment steps). Curves show \(|\delta_\Delta B(k)|\) and \(|\delta_\Delta q(k)|\), each divided by the environment's mean Euclidean squared distance \(\overline D_E\) over the tested offsets. In both environments, the \(B\) term varies little, while changes in the \(q\) term become smaller at longer offsets.}
\vspace{-15pt}
\label{fig:temporal_contrast_components}
\end{figure*}

Empirically, Fig.~\ref{fig:temporal_contrast_components} shows this behavior on 50 recorded planner executions each in AntMaze and FourRoom. We encode observations sampled every model step and use the same future goals across offsets \(k=1,\ldots,23\), comparing adjacent offsets with \(\Delta=1\). Results are averaged within each trajectory and then equally across trajectories, and both components are normalized by the mean squared distance over the tested offsets. In both environments, \(|\delta_\Delta B(k)|\) stays small, while \(|\delta_\Delta q(k)|\) decreases markedly overall as \(k\) grows. Consequently, the bound becomes small at long offsets, explaining the weak average Euclidean cost signal for distinguishing states which are distant to goal in Fig.~\ref{fig:plan_progress_fourroom}.

These observations motivate the objective of explicitly teaching the latent world model the temporal structure of progress toward the goal. 

\begin{figure*}[t]
\centering
\includegraphics[width=0.99\linewidth]{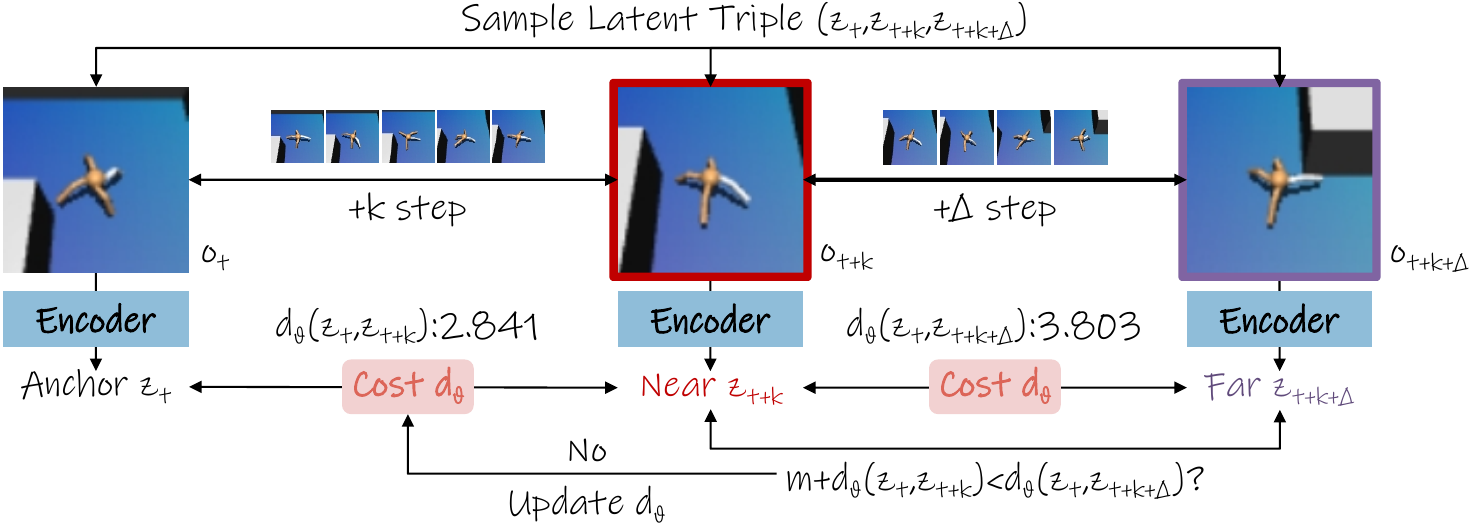}
\caption{Overview of T2G metric training. A latent triple \(({\bf z}_t,{\bf z}_{t+k},{\bf z}_{t+k+\Delta})\) is sampled from an offline trajectory, where the near and far states are separated from the anchor by \(k\) and \(k+\Delta\) steps, respectively. The three encoder blocks denote the same frozen JEPA encoder applied to the corresponding observations. T2G evaluates the learned temporal distances from the anchor to the near and far latents and optimizes the metric with a margin ranking objective, so that the near state is closer to the anchor than the far state by at least a margin \(m\). Only the metric network and its scale are updated, and the world model remains frozen. 
}
\vspace{-15pt}
\label{fig:trainframework}
\end{figure*}

\subsection{Learning A Time-to-Goal Cost}
To give the planner an explicit measure of temporal proximity, we propose T2G to learn a small mapping of the frozen JEPA representations. Let \(\theta\) denote its trainable parameters. We define the distance between two states as a scaled Euclidean distance in the learned embedding space:
\begin{equation}
	d_\theta({\bf z}_a,{\bf z}_b)=\alpha_\theta\lVert f_\theta({\bf z}_a)-f_\theta({\bf z}_b)\rVert_2,
	\label{eq:t2g_metric}
\end{equation}
where \(f_\theta\) is a small MLP and \(\alpha_\theta>0\) is a learned scale. The learned mapping reshapes the latent geometry through temporal-order supervision. The Euclidean norm preserves non-negativity, symmetry, and the triangle inequality:
\begin{equation}
	d_\theta(a,c)\le d_\theta(a,b)+d_\theta(b,c).
	\label{eq:triangle}
\end{equation}

We teach this metric to distinguish nearer and farther states using the time order of offline trajectories. The frozen encoder first maps each trajectory clip to a sequence of latent states. Within that clip, we sample three states: an \emph{anchor} state \({\bf z}_t\), with a \emph{near} state \({\bf z}_{t+k}\) and a \emph{far} state \({\bf z}_{t+k+\Delta}\). The near state occurs \(k\) model steps after the anchor, and the far state occurs another \(\Delta\) steps later. These three states form the training triple:
\begin{equation}
	(\underbrace{{\bf z}_t}_{\text{anchor}},\;
	\underbrace{{\bf z}_{t+k}}_{\text{near}},\;
	\underbrace{{\bf z}_{t+k+\Delta}}_{\text{far}}).
	\label{eq:triplet_sampling}
\end{equation}
The desired comparison is simple: the learned distance from the anchor state to the near state should be smaller than its distance to the far state. We train the metric to satisfy the near--far comparison with a margin ranking loss
\begin{equation}
	\mathcal{L}_{\mathrm{rank}}(\theta)=\mathbb{E}_{t,k,\Delta}\left[\max\left(0, m+d_\theta({\bf z}_t,{\bf z}_{t+k})-d_\theta({\bf z}_t,{\bf z}_{t+k+\Delta})\right)\right],
	\label{eq:rank_loss}
\end{equation}
where \(m>0\) is fixed during training. A triple incurs a penalty when the far state is not farther from the anchor than the near state by at least \(m\) model step. The positive scale $\alpha_\theta$ in Eq.~(\ref{eq:t2g_metric}) adjusts the magnitude of distance differences relative to $m$. We average this penalty over the minibatch and update only the metric parameters \(\theta\), while the encoder \(E\) and predictor \(F\) remain frozen. The supervision comes entirely from trajectory order, with no rewards or policy labels. Since the loss specifies relative distances rather than a target number of steps, \(d_\theta\) is a relative temporal cost. Once trained, it serves as the goal-matching cost for the frozen world model.

\subsection{Planning With the Time-to-Goal Cost}
At planning time, we replace the Euclidean goal cost of flat LeWM with the learned temporal distance. Starting from the observed latent history ending at \({\bf z}_1\) and the goal latent \({\bf z}_g\), the planner rolls out candidate action sequences \(a_{1:H}\) with the frozen predictor to obtain \({\hat {\bf z}}_{2:H+1}\). The last-\(n\) T2G cost averages the temporal distances to the goal over the final \(n\) predicted states:
\begin{equation}
\begin{aligned}
    C_{\mathrm{T2G}}^{(n)}({\hat {\bf z}}_{2:H+1})=\frac{1}{n}\sum\nolimits_{t=H-n+2}^{H+1}d_\theta(\hat {\bf z}_t,{\bf z}_g),
    \label{eq:t2g_cost_lastn}
\end{aligned}
\end{equation}
where \(1\le n\le H\). This cost focuses on the final portion of the rollout, using the learned temporal distance to evaluate how close those states are to the goal.

We also consider a \textit{dense} T2G variant that averages over the entire rollout:
\begin{equation}
\begin{aligned}
    C_{\mathrm{T2G}}({\hat {\bf z}}_{2:H+1})=\frac{1}{H}\sum\nolimits_{t=2}^{H+1}d_\theta(\hat {\bf z}_t,{\bf z}_g).
    \label{eq:t2g_cost}
    \end{aligned}
\end{equation}
The dense cost is the special case \(n=H\), incorporating intermediate predicted states as well as the final states.

Both costs can be used by sampling-based and gradient-based planners. Sampling-based planners (e.g., CEM~\cite{rubinstein2004cross}) select the lowest-cost candidate, whereas the gradient-based planner in H-JEPA minimizes the cost over actions. The planner then executes the first action block(s) and replans in a receding-horizon fashion.

\begin{table}[t]
	\centering
	\caption{Planning settings of the proposed T2G. Episode budgets are in environment steps, \(H\) is the beginning planning horizon in predicted steps, \(S\) is the number of candidate action sequences, and \(\eta\) is the AdamW learning rate. Last-\(n\) is the number of final predicted states averaged by the T2G last-\(n\) cost. Candidate counts, learning rates, and episode budgets follow the flat LeWM baseline of H-JEPA~\cite{H-JEPA}; Cube uses an initial horizon of two.}
    \vspace{5pt}
	\setlength{\tabcolsep}{7.4mm}{\begin{tabular}{lccccc}
		\toprule
		Environment & Episode budget & \(H\) & Last-$n$ & \(S\) & \(\eta\) \\
		\midrule
		FourRoomDistractors & 210 & 23 & 2& 200 & 0.1 \\
		Visual AntMaze & 350 & 23 &2& 100 & 0.3 \\
		Push-T & 90 & 15 &2& 800 & 0.05 \\
		OGBench Cube & 30 & \textit{2} &1 & 800 & 0.05 \\
		\bottomrule
	\end{tabular}}
	\label{tab:planning_settings}
\end{table}

\begin{figure*}[t]
\centering
\includegraphics[width=0.99\linewidth]{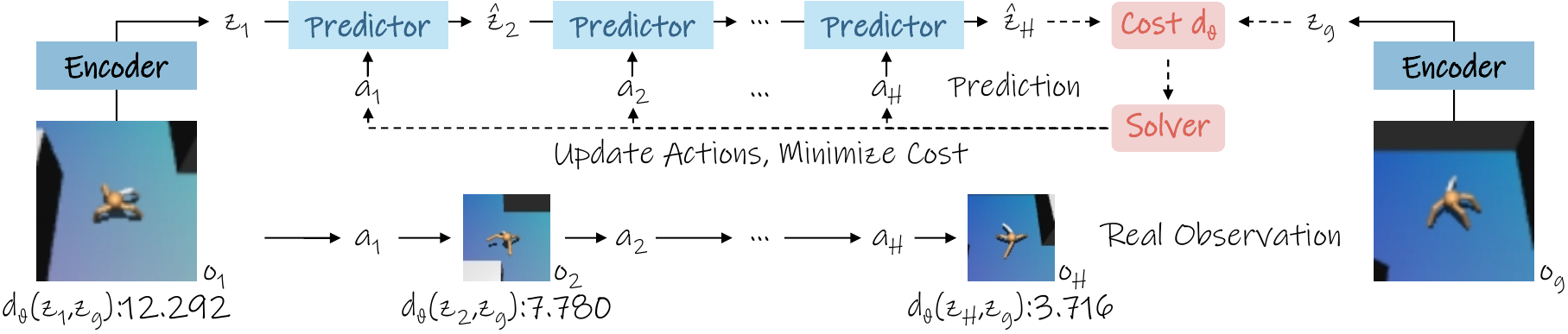}
\caption{Overview of T2G planning. The current observation \(o_1\) and goal observation \(o_g\) are encoded into latents \({\bf z}_1\) and \({\bf z}_g\) by the frozen JEPA encoder. A solver optimizes an action sequence \(a_{1:H}\), and the frozen predictor rolls out the resulting latent trajectory \({\bf z}_1\!\to\!{\bf z}_2\!\to\!\cdots\!\to\!{\bf z}_H\). Each predicted latent is scored by its learned temporal distance to the goal, \(d_\theta({\bf z}_t,{\bf z}_g)\), and the solver updates the actions to minimize the aggregated cost. The updated actions are executed and planning is repeated from new observations in a receding-horizon loop, while the encoder, predictor, and solver remain unchanged. 
}
\vspace{-15pt}
\label{fig:planframework}
\end{figure*}

\begin{table}[t]
	\centering
    \PerfDeclareRow{hjepa2}{10.79}{39.3/3.7}{47.3/2.7}{81.3/6.7}{45.3/1.3}
    \PerfDeclareRow{hjepa3}{21.58}{73.3/3.5}{60.0/5.0}{96.0/1.1}{17.3/3.5}
    \PerfDeclareRow{hjepa4}{32.37}{63.3/7.7}{55.3/4.4}{80.0/4.0}{0.7/0.7}
    \PerfDeclareRow{lewm}{--}{18.0/3.5}{32.0/6.1}{40.7/6.8}{40.0/2.3}
    \PerfDeclareRow{dense}{0.18}{48.3/2.4}{4.0/1.2}{81.3/2.9}{42.0/1.2}
    \PerfDeclareRow{lastn}{0.18}{58.7/2.4}{63.3/0.7}{78.7/2.9}{51.3/5.9}
	\caption{Planning success rate (\%) on the evaluation tasks, averaged over multiple seeds. LeWM (flat) uses the native level-1 cost; the 2-, 3-, and 4-level rows use hierarchical H-JEPA planning, and the T2G rows use the same frozen flat world model with either a full-rollout (dense) or final-state-window (last-\(n\)) temporal cost. Window sizes are given in Tab.~\ref{tab:planning_settings}. \emph{New Params} denotes the number of additional parameters in millions (M) compared to LeWM (flat). Cell shading continuously interpolates from \perfcolor{perfLow}{light red} to \perfcolor{perfHigh}{deep red} using min--max normalization independently within each column. Higher mean success rates receive darker shading; the scale is reversed for \emph{New Params}, where lower values are better. Standard errors do not affect shading. The missing entry (\texttt{--}) is unshaded and excluded from normalization.}
	\vspace{5pt}
	\renewcommand{\arraystretch}{1.15}
	\setlength{\tabcolsep}{1.7mm}{\begin{tabular}{lccccc}
		\toprule
		Method & New Params $\downarrow$ & Visual AntMaze$\uparrow$ & OGBench Cube$\uparrow$ & FourRoomDistractors$\uparrow$ & Push-T$\uparrow$ \\
		\midrule
		
		H-JEPA (2 levels)
		& \PerfCell{hjepa2}{params}
		& \PerfCell{hjepa2}{ant}
		& \PerfCell{hjepa2}{cube}
		& \PerfCell{hjepa2}{fourroom}
		& \PerfCell{hjepa2}{pusht} \\
		H-JEPA (3 levels)
		& \PerfCell{hjepa3}{params}
		& \PerfCell{hjepa3}{ant}
		& \PerfCell{hjepa3}{cube}
		& \PerfCell{hjepa3}{fourroom}
		& \PerfCell{hjepa3}{pusht} \\
		H-JEPA (4 levels)
		& \PerfCell{hjepa4}{params}
		& \PerfCell{hjepa4}{ant}
		& \PerfCell{hjepa4}{cube}
		& \PerfCell{hjepa4}{fourroom}
		& \PerfCell{hjepa4}{pusht} \\
        \midrule
        LeWM (flat)
		& \PerfCell{lewm}{params}
		& \PerfCell{lewm}{ant}
		& \PerfCell{lewm}{cube}
		& \PerfCell{lewm}{fourroom}
		& \PerfCell{lewm}{pusht} \\
		+ T2G dense
		& \PerfCell{dense}{params}
		& \PerfCell{dense}{ant}
		& \PerfCell{dense}{cube}
		& \PerfCell{dense}{fourroom}
		& \PerfCell{dense}{pusht} \\
        + T2G lastn
		& \PerfCell{lastn}{params}
		& \PerfCell{lastn}{ant}
		& \PerfCell{lastn}{cube}
		& \PerfCell{lastn}{fourroom}
		& \PerfCell{lastn}{pusht} \\
		\bottomrule
	\end{tabular}}
	\label{tab:planning_results}
\end{table}

\begin{figure*}[t]
\centering
\includegraphics[width=0.99\linewidth]{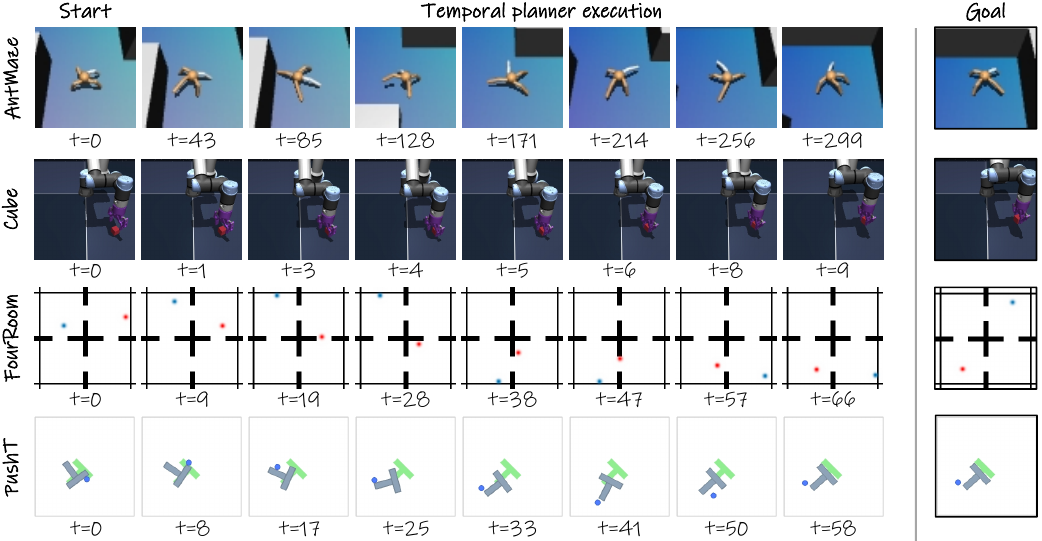}
\caption{Selected successful LeWM+T2G planner executions using our T2G cost, on one evaluation task per environment. 
}
\vspace{-5pt}
\label{fig:successvis}
\end{figure*}

\section{Experiments}

\subsection{Planning Evaluation Setup}
\textbf{Datasets and Tasks.} We evaluate T2G on four simulated benchmarks used by H-JEPA~\cite{H-JEPA}: FourRoomDistractors~\cite{FourRoom}, Visual AntMaze~\cite{park2025ogbench}, Push-T~\cite{pusht}, and OGBench Cube~\cite{park2025ogbench}. These cover navigation with distractors, quadruped maze navigation, planar manipulation, and 3D object manipulation, respectively. We follow H-JEPA's task construction. FourRoomDistractors and Push-T use held-out expert frame pairs separated by 75 environment steps, with the FourRoom start and goal located two rooms apart. Cube uses held-out 20-step trajectory segments centered on a grasp. AntMaze uses expert rollouts between maze cells at grid distance three, spanning about 82 environment steps on average, with neutral-pose start and goal states. The planner receives a goal image and additionally uses proprioceptive observations on AntMaze. We evaluate on 50 start--goal tasks per environment, shared across all methods and hierarchy depths. Success is determined by the environment's goal test within the episode budget; on AntMaze, the agent must reach within 1.0 of the goal \(xy\) position. Planner settings are fixed before the final evaluation.

\textbf{Planner and Baselines.} We compare flat LeWM with its Euclidean goal cost, flat LeWM with last-\(n\) and dense T2G costs, and H-JEPA with two, three, and four levels. All flat variants use the gradient-based planner with AdamW. On FourRoom, AntMaze, and Push-T, T2G retains the baseline planning settings and replaces the goal cost. On Cube, T2G uses a horizon of \(H=2\) model steps, replacing the baseline's initial horizon of 4. The configured last-\(n\) window is 2 predicted states on AntMaze, FourRoom, and Push-T, and 1 on Cube; the effective window is capped by the current horizon. Tab.~\ref{tab:planning_settings} summarizes the episode budgets, T2G horizons, window sizes, candidate counts \(S\), and AdamW learning rates \(\eta\). All methods share the evaluation tasks and environment-step budgets. We report planning success rate (\%) as the mean and standard error across seeds 42, 43, and 44.


\subsection{Planning Performance Across Different Environments}
Tab.~\ref{tab:planning_results} compares flat LeWM, two T2G cost variants, and H-JEPA with two to four levels. Specifically, T2G last-\(n\) improves the mean success rate over flat LeWM in all four environments: from 18.0\% to 58.7\% on AntMaze, 32.0\% to 63.3\% on Cube, 40.7\% to 78.7\% on FourRoom, and 40.0\% to 51.3\% on Push-T. Its unweighted average across environments is 63.0\%, compared with 32.7\% for flat LeWM and 53.3\%, 61.7\%, and 49.8\% for two-, three-, and four-level H-JEPA, respectively (summarize the reported means). Both T2G variants retain the flat world model and add only a small metric network.

Dense T2G also has substantial effect on several environments. It improves AntMaze, FourRoom, and Push-T to 48.3\%, 81.3\%, and 42.0\%, respectively, but reduces Cube success to 4.0\%. Restricting the cost to the final \(n\) predicted states (last-$n$ T2G) raises the corresponding means on AntMaze, Cube, and Push-T by 10.4\%, 59.3\%, and 9.3\%, while FourRoom slightly decreases by 2.6\%. Thus,  uniformly aggregating full-rollout temporal distance does not make planning consistently effective. In particular, dense aggregation encourages proximity to the goal at every predicted step, including the initial approach and grasp preparation of Cube. These necessary intermediate states may still receive large T2G distances to the goal. Penalizing them can favor actions that reduce the early cost at the expense of completing the grasp, whereas last-1 scoring allows the planner to optimize these preparatory actions for their effect on the predicted endpoint. This provides an explanation for the large advantage of terminal-state scoring on Cube. 

Relative to hierarchy, T2G last-\(n\) has the highest reported mean on Cube and Push-T, exceeding the best H-JEPA means by 3.3\% and 6.0\%. T2G substantially narrows the performance gap between flat LeWM and hierarchical planning on the navigation tasks. On FourRoom, dense T2G raises success from 40.7\% to 81.3\%, matching two-level H-JEPA. Three-level H-JEPA achieves higher success on both AntMaze and FourRoom, at 73.3\% and 96.0\%, respectively. Together, these results show that a learned temporal cost can recover substantial performance within a flat architecture and motivate combining T2G with hierarchical planning to further improve navigation.

\subsection{Temporal Guidance in AntMaze and FourRoom}
\label{sec:temporal_diagnostics}

\begin{figure*}[t]
\centering
\includegraphics[width=0.99\linewidth]{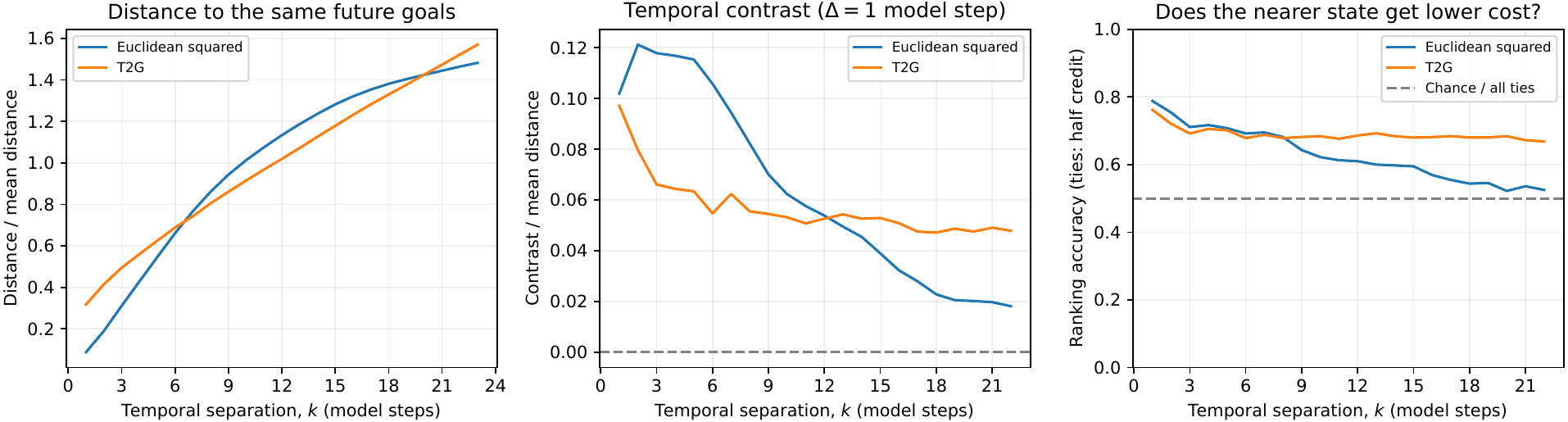}\par
\vspace{5pt}
\includegraphics[width=0.99\linewidth]{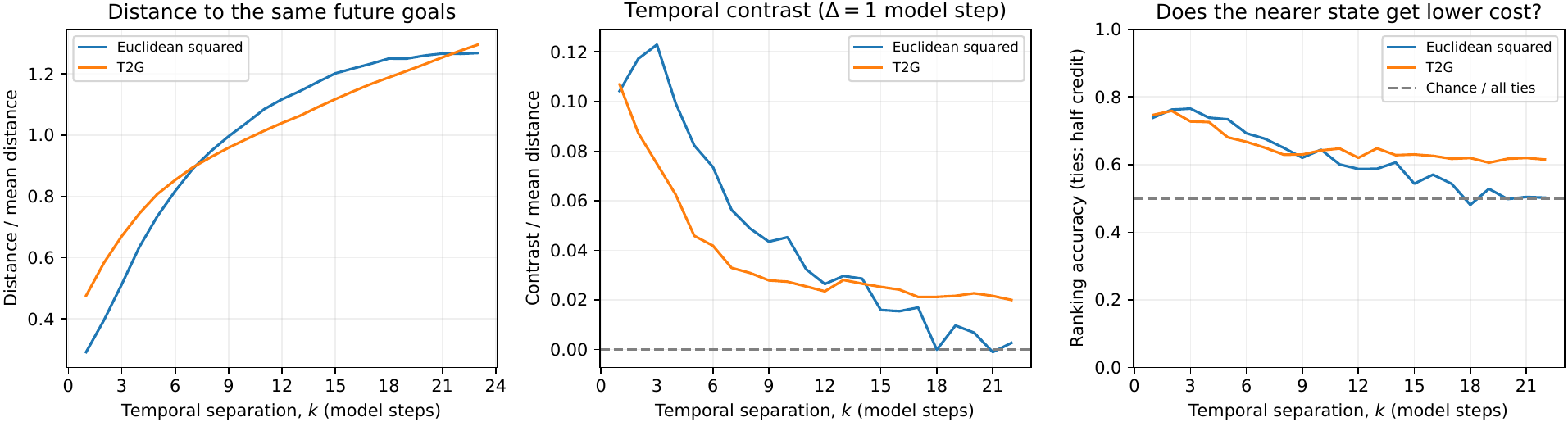}
\caption{Temporal guidance of Euclidean squared and T2G costs on recorded AntMaze (top) and FourRoom (bottom) executions. Left: mean cost to the same sampled future goals, normalized by each cost's mean over the tested offsets. Middle: normalized temporal contrast between offsets separated by one model step (five environment steps). Right: the fraction of comparisons in which the temporally nearer state receives a lower cost, with ties receiving half credit; the dashed line marks 50\%. Horizontal axes report temporal separation in environment steps. At long offsets, T2G maintains a larger positive contrast and higher ranking accuracy in both environments.}
\vspace{-15pt}
\label{fig:temporal_signal}
\end{figure*}

To examine the temporal guidance discussed in Section~\ref{sec:euclidean_cost}, Fig.~\ref{fig:temporal_signal} compares Euclidean squared and T2G costs on 50 recorded executions per environment from flat LeWM with T2G last-$2$. Both costs score the same encoded observations and sampled future goals. Within each trajectory, the goal set is held fixed across offsets \(k=1,\ldots,23\) model steps, corresponding to 5--115 environment steps on the figure axes. We compare adjacent offsets with \(\Delta=1\) model step and average first within each trajectory and then equally across trajectories. Each cost curve and its temporal contrast are divided by that cost's mean over the tested offsets, enabling comparison of their shapes on a common relative scale.

The left panels show that both costs increase overall with temporal separation. The Euclidean curves, however, become progressively flatter at long offsets, particularly in FourRoom, while T2G continues to distinguish successive offsets. The middle panels quantify this difference through the temporal contrast in Eq.~(\ref{eq:temporal_distance_contrast}). Over the last five comparisons, \(k=18,\ldots,22\), the mean normalized contrast is 0.0202 for Euclidean cost and 0.0481 for T2G on AntMaze, and 0.0036 versus 0.0214 on FourRoom. Thus, T2G preserves a stronger average cost difference between temporally nearer and farther states in both environments. The effect is especially pronounced in FourRoom, where the Euclidean contrast fluctuates around zero at the largest offsets. These observations agree with the component analysis in Fig.~\ref{fig:temporal_contrast_components}, which links weak Euclidean contrast to small changes in \(B(k)\) and \(q(k)\).

The right panels assess whether the cost also orders individual state pairs correctly: for the same goal, does the state at offset \(k\) receive a lower cost than the state at \(k+1\)? Euclidean cost provides useful ordering at short offsets, but its accuracy decreases toward 50\% as the separation grows. Over the same five long-offset comparisons, T2G raises ranking accuracy from 53.44\% to 67.69\% on AntMaze and from 50.27\% to 61.53\% on FourRoom, gains of 14.25 and 11.26 percentage points, respectively. T2G's advantage therefore emerges most clearly when the goal is temporally distant, where the original Euclidean cost becomes less discriminative.

Together, these measurements connect the temporal-contrast analysis to the ordering learned by T2G: the learned cost provides both larger average differences and more accurate preferences for temporally nearer states at long offsets. This supplies the flat planner with a clearer signal for comparing candidate futures using the frozen predictive representation.


\section{Related Works}
JEPAs learn task-agnostic world models by predicting future observations in a latent space and have been extended to action-conditioned latent planning~\cite{lecun2022path,assran2023self,bardes2024revisiting}. DINO-WM trains dynamics on frozen visual features~\cite{zhou2024dino}; PLDM learns latent dynamics with auxiliary regularization~\cite{sobal2026learning}; V-JEPA~2 attaches a predictor to a pretrained video JEPA~\cite{assran2025v}; and LeWM trains an encoder and predictor end-to-end with SIGReg~\cite{LeWM}. These reward-free models are typically paired with MPC, which scores candidate rollouts using CEM~\cite{rubinstein2004cross} or gradient-based optimization. Reward-based approaches such as Dreamer, TD-MPC2, and MuZero instead learn models for reward or value maximization~\cite{hafner2023mastering,hansen2024td,schrittwieser2020mastering}. To handle long horizons, hierarchical methods decompose planning across timescales, from options~\cite{sutton1999between} to HWM, which plans over multiple horizons in a shared latent space~\cite{zhang2026hierarchical}, and H-JEPA, which stacks JEPAs with increasing temporal strides and predicts subgoals~\cite{H-JEPA}. T2G adds no hierarchy and changes no world model; it changes only the cost used by a flat planner.

Temporal progress has also been studied through time-contrastive and predictive objectives~\cite{sermanet2018time,oord2018representation}, goal-conditioned value functions and relabeling~\cite{andrychowicz2017hindsight,ghosh2019learning}, and quasimetric RL. These methods typically learn a policy or value function and rely on interaction or reward. T2G instead learns a embedded metric from offline latents of a frozen JEPA model with a margin ranking objective, requires no reward or policy signal, and serves directly as a planning cost for sampling- or gradient-based planners. This isolates cost geometry from temporal abstraction and complements hierarchy.






\section{Conclusion}
The central finding of this work is that the cost used to evaluate predicted states plays an important role in the effectiveness of latent planning. T2G demonstrates that trajectory order provides useful supervision for shaping this cost, allowing a frozen predictive representation to offer clearer temporal guidance. The comparison between dense and last-\(n\) costs further shows that the choice of scored states matters: emphasizing the final predicted states can accommodate intermediate actions needed to complete a task. These findings make learning and aggregating goal costs a practical part of designing planners with JEPA world models.

Future work will combine T2G with hierarchical planning by learning temporal costs at multiple levels to guide both subgoal selection and action optimization. This would bring the improved temporal ordering of T2G together with the longer planning range provided by temporal abstraction. We also plan to extend the approach to physical robots, using offline robot trajectories to learn the temporal metric and evaluating it during closed-loop task execution. Such experiments will examine how the learned cost supports progress toward a goal under real visual variation, sensor noise, and contact dynamics.

\section*{Acknowledgments}
This was was supported in part by National Natural Science Foundation of China (62506067).

\bibliographystyle{unsrt}  
\bibliography{main}  

\newpage
\appendix

\section{T2G training details}
\label{app:t2g_training}

\textbf{Device.} All experiments were conducted on the same hardware environment running Ubuntu 22.04.2 LTS, with 256 GB of RAM and Intel Xeon Silver4320. A single NVIDIA A100 GPU provides the computing acceleration in experiments.

\textbf{Sampling process during training.} For each clip, let \(T_{\mathrm{clip}}\) be the number of latent states and let
\(K=\min(K_{\max},T_{\mathrm{clip}}-1)\) be the largest usable offset from the anchor. Here,
\(K_{\max}\) is the configured maximum offset, \(k_{\min}\) is the minimum
near-state offset, and \(\Delta_{\min}\) is the minimum near--far gap; all
offsets are measured in model steps. We require
\(K\ge k_{\min}+\Delta_{\min}\), so that the clip can accommodate both a near
and a far state. To sample a triple, we draw the separations first and then the
anchor:
\begin{enumerate}
	\item Draw the near-state offset \(k\) uniformly from
	\(\{k_{\min},\ldots,K-\Delta_{\min}\}\). The upper bound leaves at least
	\(\Delta_{\min}\) steps between the near and far states.
	\item Draw an additional gap \(e\) uniformly from
	\(\{\Delta_{\min},\ldots,K\}\).
	\item Set the far-state offset to \(k_{\mathrm{far}}=\min(k+e,K)\), and set
	\(\Delta=k_{\mathrm{far}}-k\). Hence \(\Delta\ge\Delta_{\min}\) and
	\(k_{\mathrm{far}}\le K\). If \(k+e>K\), the far offset is clipped to
	\(K\).
	\item Draw the anchor index \(t\) uniformly from
	\(\{0,\ldots,T_{\mathrm{clip}}-1-k_{\mathrm{far}}\}\), so that the far state remains
	inside the clip.
\end{enumerate}
The resulting triple is \((z_t,{\bf z}_{t+k},{\bf z}_{t+k+\Delta})\). Each clip in a minibatch contributes one
newly sampled triple. Clipping gives extra probability to the maximum
far-state offset, increasing exposure to long temporal separations within the training window.

\textbf{Hyper-parameter settings.} For each environment, T2G is trained on the corresponding offline training split (provided by \texttt{\href{https://huggingface.co/datasets/jepa-world-models/h-jepa/tree/main}{https://huggingface.co/datasets/jepa-world-models/h-jepa/tree/main}}) using the frozen flat LeWM encoder and predictor (provided by \texttt{\href{https://huggingface.co/jepa-world-models/h-jepa}{https://huggingface.co/jepa-world-models/h-jepa}}). The metric mapping \(f_\theta\) consists of two hidden blocks, each containing a linear layer of width 256, LayerNorm, and GELU, followed by a linear output layer of dimension 256. Dropout is set to zero. The input dimension follows the LeWM representation: 384 for AntMaze and 192 for the other environments. The positive distance scale \(\alpha_\theta\) is parameterized with softplus and initialized to 1. The computed distance is lower-bounded by \(10^{-6}\) before scaling.

All four environments use AdamW with a fixed learning rate of \(5\times10^{-5}\), weight decay \(10^{-3}\), batch size 128, and 5 training epochs. We minimize the margin ranking loss with \(m=1\) and clip the gradient norm of the metric parameters to 1.0. Training clips are shuffled at each epoch, with incomplete final minibatches dropped. Each clip contributes one newly sampled triple per update, giving 128 triples per minibatch. Only the metric mapping and its scale are optimized; the world model is kept in evaluation mode. Each metric uses the original LeWM checkpoint.

Tab.~\ref{tab:t2g_training_sampling} lists the environment-specific input and sampling settings. Observations are sampled every five environment steps, so one model step corresponds to five environment steps in all four environments. We set \(k_{\min}=1\) and \(\Delta_{\min}=2\) throughout. The latter is the minimum near--far gap, and the actual gap \(\Delta\) varies across sampled triples. The maximum training offset \(K_{\max}\) specifies the sampling range and is distinct from the planning horizon \(H\).

\begin{table}[t]
\centering
\caption{Input and sampling hyper-parameters for T2G training. Image size is in pixels. Clip length \(T_{\mathrm{clip}}\) counts latent states, and \(K_{\max}\) is measured in model steps. All environments use \(k_{\min}=1\), \(\Delta_{\min}=2\), and five environment steps per model step.}
\label{tab:t2g_training_sampling}
\begin{tabular}{lcccc}
\toprule
Environment & Latent dimension \(D\) & Image size & \(T_{\mathrm{clip}}\) & \(K_{\max}\) \\
\midrule
AntMaze & 384 & \(64\times64\) & 24 & 23 \\
FourRoom & 192 & \(224\times224\) & 24 & 23 \\
Cube & 192 & \(224\times224\) & 16 & 15 \\
Push-T & 192 & \(224\times224\) & 16 & 15 \\
\bottomrule
\end{tabular}
\end{table}

\end{document}